\documentclass[10pt,twocolumn,letterpaper]{article}

\usepackage{cvpr}              

\usepackage{amsmath}
\usepackage{amssymb}
\usepackage{tipa}
\usepackage{colortbl}
\usepackage{pifont} 
\usepackage{makecell}
\usepackage[xcdraw,table]{xcolor}
\usepackage{tikz}     
\usetikzlibrary{external}
\usepackage{float} 
\usepackage{caption}
\usepackage{subcaption}
\usepackage{booktabs}
\usepackage{lipsum}
\usepackage{graphicx}
\usepackage{multirow}
\usepackage{array}
\usepackage{algorithm}
\usepackage{algorithmic}
\usepackage{setspace}

\renewcommand{\arraystretch}{1.2}
\usepackage{xspace}
\newcommand{\mymethod}{ANoCo\xspace}

\definecolor{Gray}{gray}{0.95}
\newcommand{\gI}{\mathcal{I}}
\newcommand{\gA}{\mathcal{A}}
\newcommand{\gF}{\mathcal{F}}

\definecolor{cvprblue}{rgb}{0.21,0.49,0.74}
\usepackage[pagebackref,breaklinks,colorlinks,allcolors=cvprblue]{hyperref}

\def\paperID{15195} 
\def\confName{CVPR}
\def\confYear{2026}

\title{Anomaly as Non-Conformity\\ via Training-Free Graph Laplacian Energy Minimization}

\author{
Jungwook Seo$^{1}$ \quad
Minjeong Kim$^{1}$ \quad
Younkwan Lee$^{3}$ \quad
Seungho Shin$^{3}$ \quad
Sungyong Baik$^{1,2\dagger}$\\
$^{1}$Dept. of Artificial Intelligence, Hanyang University\\
$^{2}$Dept. of Data Science, Hanyang University\\
$^{3}$Global Technology Research, Samsung Electronics\\
{\tt\small \{zungwooker,mjkim0720,dsybaik\}@hanyang.ac.kr}\\
{\tt\small \{youn720.lee,sh524.shin\}@samsung.com}
}

\begin{document}
\maketitle
\footnotetext[2]{Corresponding author.}
\begin{abstract}
Detecting subtle visual anomalies in images remains challenging, particularly when only normal samples are available \`a priori.
Such unsupervised anomaly detection is typically solved by measuring feature similarity of a query patch to a memory of normal patches.
However, similarity alone does not reveal how strongly a query patch violates the structure of the normal feature manifold.
We propose a training-free Laplacian graph energy optimization formulation, named \textbf{ANoCo} that scores \textbf{A}nomaly by the cost of \textbf{No}n-\textbf{Co}nformity of a query patch to align with a fixed normal manifold.
For each query patch, we construct a bipartite query\textrightarrow normal graph weighted by cosine affinity, explicitly removing query–query and normal–normal edges to prevent evidence dilution. 
We formulate anomaly scoring as a convex Laplacian energy with anchored normal nodes, and solve in closed form. 
In particular, we do not use the optimized features themselves—the anomaly score is the magnitude of the update required to satisfy normality constraints, reframing the graph Laplacian as a non-conformity operator rather than a smoothing prior. 
The proposed method introduces no learnable parameters, message passing, or sampling, and has complexity comparable to a single linear solve. 
Across standard benchmarks, it delivers strong image-level AUROC, stable localization maps, and improved robustness over prior methods, demonstrating the effectiveness of using optimization-induced feature drift as anomaly measure.
\end{abstract}    
\begin{figure}[t]
    \centering
    \includegraphics[width=\columnwidth]{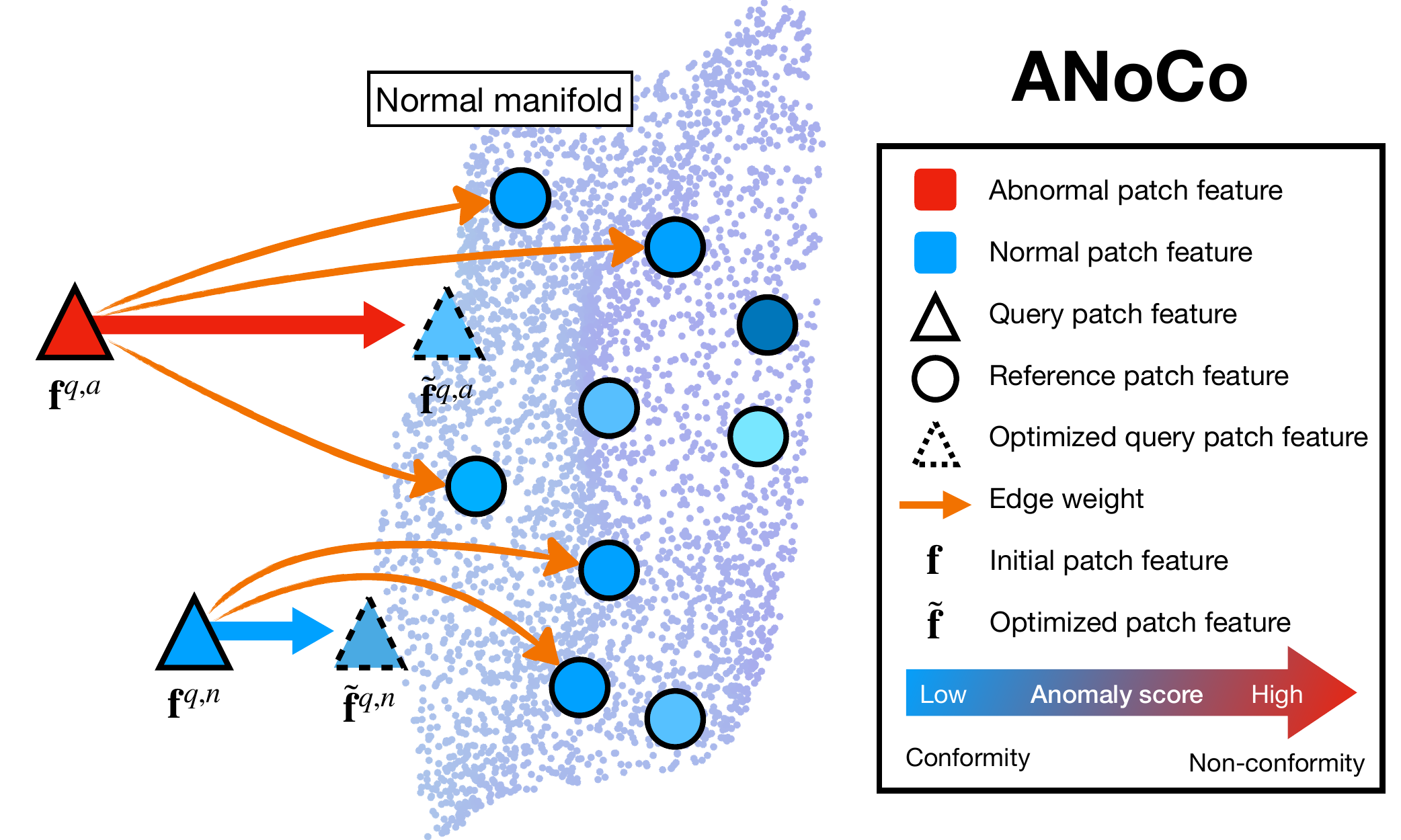}
    \caption{\textbf{Conceptual illustration of ANoCo.}
ANoCo adjusts query patch features through an anchored Laplacian energy optimization, producing features that conform to the normal manifold.  
The magnitude of the optimization-induced feature update serves as the anomaly score, providing a simple and training-free measure of non-conformity.
}
\label{fig:cr_teaser}
\vspace{-0.4cm}
\end{figure}

\section{Introduction}
\label{sec:intro}
Industrial anomaly detection aims to identify defective samples (anomalies) using only normal (defect-free) product images, since anomalies are rare, diverse, and prohibitively expensive to annotate for supervision.
For fast adaptation of anomaly detection to new products, there has been growing interests in a few-shot normal regime, where only a handful of normal exemplars are available per object or category. 
Thus, few-shot industrial anomaly detection poses as a challenging task, which requires the capability of modeling the behavior of normality with high fidelity, while remaining sensitive to fine-grained deviations to detect anomalies.
Consequently, recent research has shifted toward training-free and retrieval-based paradigms, leveraging frozen feature extractors and exemplar comparison~\cite{spade, padim, patchcore, winclip, graphcore}.

Recent research has shin representation learning enable highly effective anomaly detectors through nearest-neighbor search in deep feature space, where abnormal patches are identified by poor similarity to the closest features in a normal memory bank~\cite{patchcore, padim, spade}.
However, they score anomaly strictly by independent similarity to neighbors, implicitly assuming that the presence of any similar patch implies normality.
This assumption breaks down when a query is individually similar to multiple retrieved normal patches, yet incompatible with the underlying structure they collectively form.
In few-shot industrial settings, where normal exemplars capture multiple modes of appearance (e.g., lighting, texture, manufacturing tolerances), $k$-NN may retrieve a plausible neighbor set that is not mutually coherent, allowing anomalous patches to be mistakenly accepted due to fragmented or conflicting local matches.
This reveals a core limitation of independent similarity scoring—it cannot detect when a query is locally plausible but globally inconsistent with the structure of the retrieved normal manifold.

Meanwhile, graph-based approaches have recently been explored to model relationships beyond isolated comparisons~\cite{graphcore, kag_prompt}. 
However, most adopt message passing, Laplacian smoothing, or homophily assumptions, implicitly enforcing agreement between connected nodes~\cite{deeper, zhou2020graph, zhu2020beyond}. 
While effective for representation learning and semi-supervised classification~\cite{deeper, kipf2016semi, feng2020graph}, these mechanisms are counterproductive for anomaly detection, as they blur deviations via neighborhood consensus and propagate evidence across queries or reference nodes~\cite{tang2022rethinking, zhao2020error, kim2022graph}. 
This exposes a core contradiction: methods that encourage smoothness on the graph are likely to suppress exactly the signals required to detect anomalies. 
 
In this work, we shift the question from ``Is this patch similar to normal?'' to ``How difficult is it to make this patch conform to normal?''.
In other words, we argue that the key signal for anomaly is non-conformity to the retrieved normals as a structured set.
We realize \textbf{A}nomaly as \textbf{No}n-\textbf{Co}nformity via a new framework (\textbf{\mymethod{}}), which is a training-free, energy-based framework that defines anomaly as the cost required to conform to a normal manifold. 
Notably, our proposed method retrieves similar normal patches prior to graph construction, but retrieval is not used to score anomaly—it is used only to build a reference normal manifold against which all query patches are optimized.
For each query patch, retrieved normal references are then anchored (frozen) via strong penalties, forming a stable proxy for the normal manifold.
We then construct a graph between all query patches and the retrieved normal patches, and solve a single convex anchored graph Laplacian energy minimization over all query patches jointly. 
The magnitude of change in features after optimization reveals how much each query patch must shift to align with the normal manifold, producing a direct measure of non-conformity that cannot be affected by supporting evidence from other abnormal patches in the same test image.

\mymethod{} repurposes the graph Laplacian from a smoothing operator~\cite{belkin2006manifold} into a non-conformity operator that measures resistance to normality, conceptually aligning anomaly with optimization cost rather than feature distance or density.
\mymethod{} yields multiple practical benefits: (i) anomaly scores have a clear physical meaning—the energy needed to fit the normal manifold; (ii) the solution is closed-form, convex and training-free; and (iii) by avoiding smoothing and message passing, the method preserves anomaly evidence.
Experimental results display the strong performance of \mymethod{}, demonstrating the effectiveness of our repurposed graph Laplacian in measuring non-conformity of query images to normal manifold and detecting anomalies.
\vspace{0.2cm}
\section{Related work}
\label{sec:related_work}

\noindent\textbf{Memory-based and nearest-neighbor anomaly detection.} 
Deep feature memory banks coupled with nearest-neighbor matching have become a strong baseline for anomaly detection. 
SPADE~\cite{spade} and PaDiM~\cite{padim} demonstrate the effectiveness of patch-level representation matching for localization. 
PatchCore~\cite{patchcore} further shows that core-set based memory retrieval achieves state-of-the-art performance at scale. 
Subsequent work expands upon adaptive neighborhood aggregation and improved similarity metrics~\cite{pni, k_nnn, crad}. 
However, these approaches score each query independently, without explicitly reasoning about collective compatibility with a normal manifold.
By contrast, our method maintains the memory-based paradigm but evaluates the joint conformity of a query to normal manifold through a graph-induced energy, rather than via isolated nearest-neighbor matching.

\noindent\textbf{Graph models for anomaly detection.}
Graph-based anomaly detection has been explored under several settings:
(i) Graph Neural Network (GNN)-based representation learning, where message passing is used to learn normality signals~\cite{ding2019deep, tang2022rethinking};
(ii) graph-level anomaly scoring using structural inconsistency or node influence~\cite{zhao2020error, tang2022rethinking}; and
(iii) semi-supervised label propagation on graphs, where normal labels diffuse via graph connectivity~\cite{belkin2006manifold, kipf2016semi}.
These approaches fundamentally rely on homophily~\cite{zhu2020beyond} or topology-based regularization, assuming that connected nodes should agree or share similar behavior. 
By contrast, anomaly detection in feature space exhibits the opposite requirement~\cite{kim2022graph}: abnormal regions must remain in disagreement with the normal structure. 
Our method avoids query–query or reference–reference interactions, constructing a bipartite compatibility graph that prevents anomalies from self-reinforcing through neighborhood consensus.

\begin{figure*}[t!]
    \centering
    \includegraphics[width=\textwidth]{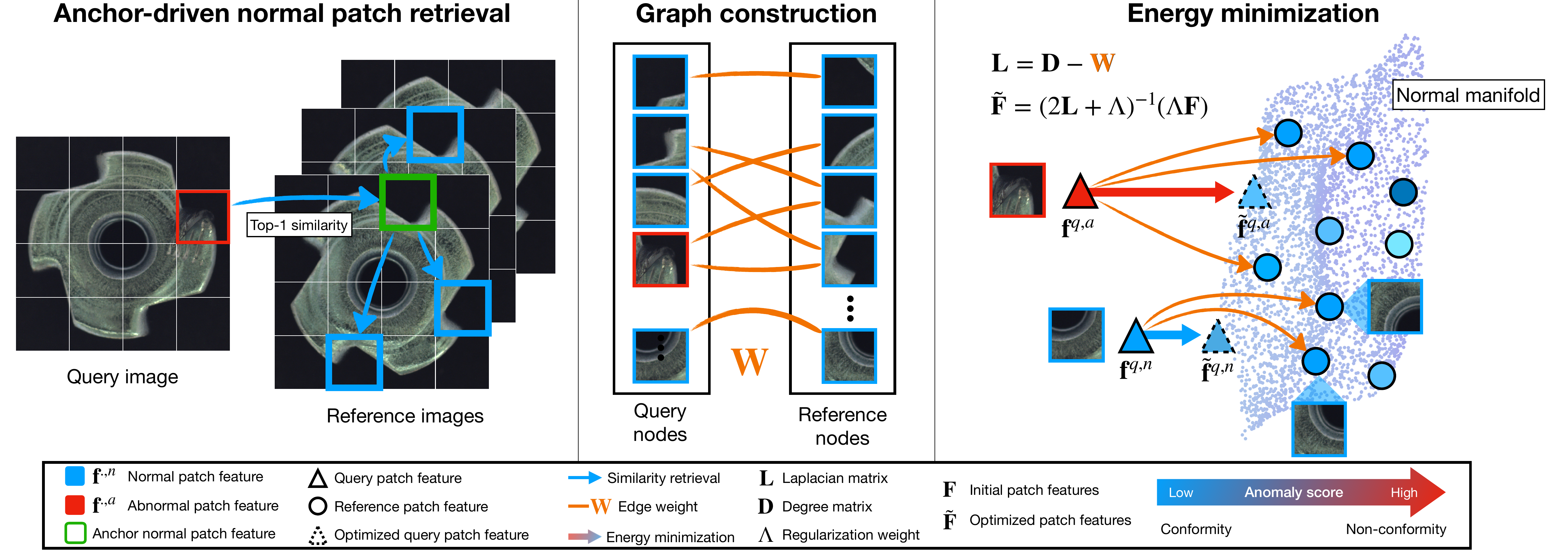}
    \vspace{-0.6cm}
    
    \caption{\textbf{Overview of \mymethod{}.}
Given a query image, similar normal reference patches are first retrieved for each query patch to form a compact normal support set. 
A bipartite graph is then constructed between query and reference patches based on their feature affinities. 
Anchored graph Laplacian optimization produces adjusted query features $\bold{\tilde{F}}_q$ that conform to the normal manifold. 
The discrepancy between original and optimized features is converted into patch-wise anomaly energies, yielding both the anomaly map and the image-level anomaly score.
}
\label{fig:cr_overview}
\vspace{-0.3cm}
\end{figure*}

\noindent\textbf{Laplacian regularization and graph smoothing.} 
Graph Laplacians are widely used for smoothing, manifold regularization, and semi-supervised learning~\cite{belkin2006manifold}, where objectives encourage neighboring nodes to share similar states. 
Several anomaly detection methods adopt Laplacian regularization for embedding refinement or diffusion~\cite{verdoja2020graph, tang2022rethinking, akoglu2015graph}, but smoothing reduces feature variance across the graph, which can suppress anomalous deviation. 
Our work reinterprets the Laplacian differently: instead of a smoothness prior, it becomes a non-conformity operator that measures the cost of reconciling a query with a frozen normal manifold. 
Furthermore, unlike prior methods that update all nodes, we anchor reference nodes with high-energy penalties, ensuring only queries move during optimization.

\noindent\textbf{Energy and optimization-based anomaly detection.}
Optimization objectives have been used to characterize anomaly through reconstruction cost, robust subspace fitting, or coding complexity~\cite{matsubara2020deep}. 
Other works interpret anomaly as the residual of a constraint-satisfying optimization process~\cite{peng2018anomalous}. 
Our \mymethod{} shares a similar energy-based spirit, but differs substantially in formulation: we do not reconstruct inputs or learn a global model. 
Instead, our proposed framework solves a Laplacian-constrained projection onto a fixed normal feature memory, and use the induced query update magnitude itself as the anomaly score—an interpretation of graph energy as manifold incompatibility, rather than reconstruction error or density violation.
\section{Proposed method}
\label{sec:proposed_method}
In this work, we reinterpret anomaly detection as measuring the non-conformity of a query patch to a normal manifold derived from a few reference images. 
From this perspective, the anomaly score is the energy required to align a query with the normal manifold.
We realize this idea by formulating anomaly detection as an anchored Laplacian energy minimization problem on a bipartite query–reference graph.
Traditional graph-Laplacian formulations enforce smoothness and drive nodes toward local consensus—an undesirable behavior for anomaly detection, where collapsing distinctions destroys the very signal of interest. 
Our framework, \textbf{A}nomaly as \textbf{No}n-\textbf{Co}nformity (\textbf{\mymethod{}}), deliberately breaks this convention.
As illustrated in Figure~\ref{fig:cr_overview}, \textbf{\mymethod{}} proceeds in three stages: \textbf{(1)} anchor-driven retrieval of normal patches to define a rigid, high-stiffness manifold with high anchor weight (Section~\ref{sec:retrieval}); \textbf{(2)} construction of a query–reference graph linking each query patch to its retrieved normals (Section~\ref{sec:graphConstruction}); and \textbf{(3)} an anchored Laplacian energy minimization (Section~\ref{sec:anchoredEnergy}), followed by anomaly scoring (Section~\ref{sec:anomaly_score}).
Notably, we do not treat the optimized query features as predictions; the required deformation itself—the feature drift induced by solving the anchored Laplacian—is the anomaly evidence, quantifying the conflict between the query and the normal manifold.  
\mymethod{} repurposes the Laplacian from a smoothing operator into a manifold-conformity operator, where optimization cost directly encodes non-conformity to normality.

\subsection{Problem formulation and motivation}
\label{sec:formulation}
Given a small set of normal reference images 
$\gI_r = \{ I^{(1)}, \dots, I^{(K)} \}$, where $K \le 4$ in the few-shot setting, and a query image $I_q$ with unknown label (either normal or abnormal), the goal is to compute an image-level anomaly score $S(I_q)$ and a dense anomaly map $\gA_q$ that highlights anomalous regions.

Let $\gF_r = \{\mathbf{f}^{r}_j \}_{j=1}^{N_r}$ and $\gF_q = \{\mathbf{f}^{q}_i \}_{i=1}^{N_q}$ denote patch features extracted via a feature extractor (e.g., ViT~\cite{vit}, DINO~\cite{dinov3}) from the reference and query images, respectively, where each feature vector corresponds to a local patch arranged on a regular two-dimensional grid.

Our core objective is to find a set of optimized features $\mathcal{\tilde{F}}_q = \{\tilde{\mathbf{f}}^{q}_i \}_{i=1}^{N_q}$ that best conform to the normal manifold, and use the resulting feature drift as evidence of anomaly.

\subsection{Anchor-driven normal patch retrieval}
\label{sec:retrieval}
\label{sec:normalRetrieval}
Given query features $\mathcal{F}_q = \{\mathbf{f}^{q}_i\}_{i=1}^{N_q}$ and the reference feature pool $\mathcal{F}_r = \{\mathbf{f}^{r}_j\}_{j=1}^{N_r}$, we first retrieve, for each query patch, a compact and reliable set of normal reference patches that locally represent the normal manifold.

\noindent\textbf{Initial similarity and anchor selection.}
For each pair of query feature $\mathbf{f}^{q}_i$ and reference feature $\mathbf{f}^{r}_j$, we compute cosine similarity $s_{ij} = \cos(\mathbf{f}^{q}_i, \mathbf{f}^{r}_j)$.  
The cosine similarities are then used to retrieve its most similar reference normal patch $\mathbf{f}^{r}_{j^\star(i)}$ for each $i$-th query patch, where $j^\star(i) = \arg\max_j s_{ij}$.
We call such retrieved reference normal patch an anchor normal patch.

\noindent\textbf{Anchor-driven retrieval.}
A na\"ive retrieval strategy would retrieve normal patches according to the similarity with a query patch, $s_{ij}$.
However, this traditional strategy introduces neighbors that are locally similar to the query but incompatible with the dominant normal pattern represented by the anchor $\mathbf{f}^{r}_{j^\star(i)}$ with the corresponding query-similarity score $s_{ij^\star(i)} $.
To obtain a more coherent normal manifold, we further retrieve normal patches that are also compatible with the anchor normal patch feature $\mathbf{f}^{r}_{j^\star(i)}$.
To this end, we first compute the similarity between an anchor normal patch and other normal patches, $a_{ij} = \cos(\mathbf{f}^{r}_{j^\star(i)}, \mathbf{f}^{r}_j)$.
Then, we sort references in descending order of $s_{ij}$, followed by retaining the longest prefix for which $a_{ij} > s_{ij^\star(i)}$, yielding a sparse query-specific neighbor set
\begin{equation}
\mathcal{N}(i) = \{\, j \,\mid\, a_{ij} > s_{ij^\star(i)} \,\text{within the sorted prefix}\,\}.
\end{equation}
Only these anchor-consistent neighbors are considered in the subsequent graph.

\subsection{Graph construction}
\label{sec:graphConstruction}
\noindent\textbf{Norm-based compatibility.}
Using the anchor-consistent neighbor sets $\{\mathcal{N}(i)\}$ in Section~\ref{sec:normalRetrieval}, we construct a sparse bipartite graph that connects each query patch only to the small subset of reference patches that are consistent with its anchor.
Noting that the cosine similarity removes the magnitude information, we prioritize feature pairs with similar magnitude, $\alpha_{ij} = \frac{2\lVert\mathbf{f}^{q}_i\rVert _2 \lVert \mathbf{f}^{r}_j\rVert _2}{\lVert\mathbf{f}^{q}_i\rVert_2 + \lVert\mathbf{f}^{r}_j\rVert_2}$, which is high when norms are compatible and small otherwise.
Hence, for each query patch $i$ and each $j \in \mathcal{N}(i)$, we define a query-reference edge
\begin{equation}
w^{\mathrm{QR}}_{ij} = s_{ij}\,\alpha_{ij},
\end{equation}
where $s_{ij}$ is the cosine similarity and $\alpha_{ij}$ is the norm compatibility factor. 
For all $j \notin \mathcal{N}(i)$, we set $w^{\mathrm{QR}}_{ij}=0$.

\noindent\textbf{Compact bipartite structure.}
Since each query patch retains only few anchor-consistent neighbors, the resulting adjacency matrix is sparse.  
Reference patches that are not selected by any query (i.e., not appearing in any $\mathcal{N}(i)$) are removed, producing a compact graph involving only query nodes and actively connected reference nodes.

\noindent\textbf{Laplacian construction.}
Let $\mathbf{W}$ denote the final sparse adjacency matrix over the combined node set $\mathcal{V}_q \cup \mathcal{V}_r$.
We compute the degree matrix $\mathbf{D}$ in the standard way and form the
graph Laplacian
\begin{equation}
\mathbf{L} = \mathbf{D} - \mathbf{W}.
\end{equation}
Because edges exist only between queries and reference nodes, the graph encodes only query-to-normal relationships and avoids oversmoothing within the query set or within the reference set. 
This bipartite Laplacian serves as the foundation for the anchored energy minimization described next.

\subsection{Block structure of features and Laplacian}
\label{sec:blockStructure}
For the sake of clarity, we briefly summarize how the feature vector and Laplacian are organized under our bipartite graph.

\noindent\textbf{Stacked feature vector.}
Only the query features are optimized, while the reference features remain fixed. We therefore write the global feature vector as $\tilde{\mathbf{F}} = [\,\tilde{\mathbf{F}}_q;\, \mathbf{F}_r\,]$, where $\tilde{\mathbf{F}}_q \in \mathbb{R}^{N_q \times d}$ contains the optimized query features and $\mathbf{F}_r \in \mathbb{R}^{N_r \times d}$ contains the fixed reference features.

\noindent\textbf{Bipartite Laplacian.}
Because the graph contains only query--reference edges, the Laplacian naturally decomposes into four blocks:
\begin{equation}
\mathbf{L} 
=
\begin{bmatrix}
\mathbf{L}_{qq} & \mathbf{L}_{qr} \\
\mathbf{L}_{rq} & \mathbf{L}_{rr}
\end{bmatrix}
=
\begin{bmatrix}
\mathbf{D}_q & -\mathbf{W}_{qr} \\
-\mathbf{W}_{qr}^{\top} & \mathbf{D}_r
\end{bmatrix}.
\end{equation}

\noindent\textbf{Diagonal query block.}
Since no query--query edges exist, $\mathbf{L}_{qq}$ is strictly diagonal, with $(\mathbf{L}_{qq})_{ii}=d^{(q)}_i$ where $d^{(q)}_i=\sum_{j\in\mathcal{N}(i)} w^{\mathrm{QR}}_{ij}$.
This diagonal structure ensures that the anchored update for each query patch decouples from all others, enabling a closed-form solution using only element-wise operations.

\subsection{Anchored Laplacian energy minimization}
\label{sec:anchoredEnergy}
Given the bipartite query--reference graph constructed in  Section~\ref{sec:graphConstruction}, we seek to quantify non-conformity (i.e., how strongly each query patch must deform to agree with the normal manifold).
To this end, we introduce an optimization variable $\tilde{\mathbf{f}}^{q}_i$ for each query patch, while reference features $\mathbf{f}^{r}_j$ remain fixed.

\noindent\textbf{Stabilizing query features.}
To prevent the refined query patches from drifting excessively away from their initial representations, we impose an $\ell_2$ regularization term:
\begin{equation}
E_{\text{reg}}
=
\sum_{i=1}^{N_q}
\lambda_i \,\left\Vert
\tilde{\mathbf{f}}^{q}_i - \mathbf{f}^{q}_i
\right\Vert_2^2.
\end{equation}
This regularization encourages each updated query feature to remain close to its original counterpart, preventing degenerate updates.
For simplicity, we use a shared weight $\lambda_i$ for all queries.

\noindent\textbf{Laplacian conformity to normal manifold.}
To enforce agreement with the reference manifold, we couple each query feature to its retrieved neighbors using the bipartite graph Laplacian.
Let $\mathbf{L}$ be the Laplacian from Section~\ref{sec:graphConstruction} and let $\tilde{\mathbf{F}}$ denote the stacked features of both query and reference nodes, where reference entries are clamped to their fixed features.
Then, the manifold conformity energy is
\begin{equation}
E_{\text{lap}}
=
\tilde{\mathbf{F}}^{\!\top}\,\mathbf{L}\,\tilde{\mathbf{F}},
\end{equation}
which penalizes disagreement along edges: query features connected to similar reference patches incur lower cost when they move toward those reference
features.

\noindent\textbf{Total anchored Laplacian energy.}
The full objective combines identity anchoring and manifold conformity:
\begin{align}
E(\tilde{\mathbf{F}})
&=
E_{\text{lap}}
+
E_{\text{reg}}\nonumber\\
&=\tilde{\mathbf{F}}^{\!\top}\,\mathbf{L}\,\tilde{\mathbf{F}}
+\sum_{i=1}^{N_q}
\lambda_i \,\left\Vert
\tilde{\mathbf{f}}^{q}_i - \mathbf{f}^{q}_i
\right\Vert_2^2.
\end{align}
Since the reference features are fixed, the optimization reduces to solving a quadratic program over the query features alone.  The resulting system is strictly convex and admits a closed-form solution obtained from a sparse linear system.

\noindent\textbf{Closed-form solver}
The anchored energy in Section~\ref{sec:anchoredEnergy} is a strictly convex quadratic function of the query optimization variables $\tilde{\mathbf{f}}^{q}_i$, since all reference features remain fixed.  
Expanding the energy using the Laplacian blocks from Section~\ref{sec:blockStructure} gives the linear system
\begin{equation}
\bigl(\mathbf{L}_{qq}+\mathbf{\Lambda}_q\bigr)\,\tilde{\mathbf{F}}_q = \mathbf{\Lambda}_q\,\mathbf{F}_q - \mathbf{L}_{qr}\mathbf{F}_r,
\end{equation}
where $\mathbf{\Lambda}_q$ is the diagonal matrix of query anchor weights $\lambda_i > 0$.

Since $\mathbf{L}_{qq}=\mathbf{D}_q$ is diagonal and $\mathbf{\Lambda}_q$ has strictly positive diagonal entries,
$\mathbf{L}_{qq}+\mathbf{\Lambda}_q$ is diagonal and trivially invertible:
\begin{equation}
\tilde{\mathbf{F}}_q = \bigl(\mathbf{L}_{qq}+\mathbf{\Lambda}_q\bigr)^{-1} \bigl(\mathbf{\Lambda}_q\mathbf{F}_q - \mathbf{L}_{qr}\mathbf{F}_r\bigr).
\end{equation}
Thus each query patch solves its update independently via element-wise division.

Because $\mathbf{L}_{qq}$ is diagonal and the anchors form another diagonal matrix, the system $(\mathbf{L}_{qq}+\mathbf{\Lambda}_q)\,\tilde{\mathbf{F}}_q = \mathbf{\Lambda}_q\,\mathbf{F}_q - \mathbf{L}_{qr}\mathbf{F}_r$ reduces to independent per-patch updates. 
Each query patch requires only a single sparse query--reference aggregation (via $\mathbf{L}_{qr}\mathbf{F}_r$) followed by element-wise division by the diagonal entries of $\mathbf{L}_{qq}+\mathbf{\Lambda}_q$.
Thus no iterative optimization, no large linear solves, and no message passing steps are needed. The entire update is a closed-form, parallelizable $O(N_q\,d)$ operation with memory proportional to the number of retrieved neighbors.

\subsection{Non-conformity as anomaly evidence}
\label{sec:anomaly_score}
In this work, we do not interpret optimized features $\tilde{\mathbf{f}}^{q}_i$ as predictions.  
Instead, the \emph{required change} is considered as non-conformity, or more importantly, the anomaly signal:
\begin{equation}
\label{eq:10}
E_i = \left\Vert\tilde{\mathbf{f}}^{q}_i - \mathbf{f}^{q}_i\right\Vert_2^2 
\,\bigl(1 - \cos(\tilde{\mathbf{f}}^{q}_i, \mathbf{f}^{q}_i)\bigr).
\end{equation}
These patchwise energies form the dense anomaly map, and the image-level score $S(I_q)$ is computed via max-pooling across patches.

\section{Experiments}
\label{sec:experiments}
\subsection{Experimental settings}
\noindent\textbf{Datasets.}
We conduct experiments on two widely used industrial anomaly detection benchmarks: MVTec-AD~\cite{mvtecad} and VisA~\cite{visa}. 
MVTec-AD contains 15 categories (10 objects and 5 textures) with 3,629 normal training images and 1,725 test images, including 467 normal and 1,258 anomalous samples, all accompanied by pixel-level ground-truth annotations. 
VisA comprises 12 high-resolution object categories with 8,659 normal training images and 2,162 test images (962 normal and 1,200 anomalous), also providing pixel-level masks for all anomalous samples.
These two benchmarks cover diverse defect types and visual complexities, offering a comprehensive evaluation of both image-level and pixel-level anomaly detection.

\noindent\textbf{Evaluation metrics.}
Following standard protocols in industrial anomaly detection, we evaluate performance using six comprehensive metrics. 
We report AUROC, AUPR, and F1-MAX for image-level detection, and AUROC, PRO, and F1-MAX for pixel-level localization. 
This multi-faceted approach ensures a robust assessment of both global discriminative power and spatial localization performance across diverse anomaly types.

\begin{table*}[t]
\centering
\caption{\textbf{Comparison of anomaly detection and localization performance on the MVTec-AD and VisA datasets under different few-shot settings.} The highest score in each column is shown in \textbf{bold}, and the second highest is marked with \underline{underline}.}
\label{tab:cr_main}
\setlength{\tabcolsep}{2.5pt}
\renewcommand{\arraystretch}{1.15}
\resizebox{\textwidth}{!}{
\begin{tabular}{c c c ccc ccc ccc ccc}
\toprule
\multirow{3}{*}{\textbf{\# refs}} & \multirow{3}{*}{\textbf{Method}} & \multirow{3}{*}{\textbf{Venue}}
& \multicolumn{6}{c}{\textbf{MVTec-AD}} & \multicolumn{6}{c}{\textbf{VisA}} \\
\cmidrule(lr){4-9} \cmidrule(lr){10-15}
& & & \multicolumn{3}{c}{Image-level} & \multicolumn{3}{c}{Pixel-level} & \multicolumn{3}{c}{Image-level} & \multicolumn{3}{c}{Pixel-level} \\
\cmidrule(lr){4-6} \cmidrule(lr){7-9} \cmidrule(lr){10-12} \cmidrule(lr){13-15}
& & & AUROC & AUPR & F1-MAX & AUROC & PRO & F1-MAX & AUROC & AUPR & F1-MAX & AUROC & PRO & F1-MAX \\
\midrule

\multirow{7}{*}{1-shot}
& SPADE~\cite{spade} & arXiv 2020
& 81.0 & 90.6 & 90.3 & 91.2 & 83.9 & 42.4 & 81.7 & 83.4 & 82.1 & 96.2 & 85.7 & 40.5 \\
& PatchCore~\cite{patchcore} & CVPR 2022
& 83.4 & 92.2 & 90.5 & 92.0 & 79.7 & 58.4 & 81.6 & 84.8 & 82.5 & 96.1 & 82.6 & 41.0 \\
& WinCLIP~\cite{winclip} & CVPR 2023
& 93.1 & 96.5 & 93.7 & 95.2 & 87.1 & 55.9 & 83.8 & 85.1 & 83.1 & 96.4 & 85.1 & 41.3 \\
& PromptAD~\cite{promptad} & CVPR 2024
& 94.6 & - & - & 95.9 & 87.9 & - & 86.9 & - & - & 96.7 & 85.8 & - \\
& KAG-Prompt~\cite{kag_prompt} & AAAI 2025
& 95.8 & 98.1 & - & 96.2 & 90.8 & - & \underline{91.6} & \underline{93.2} & - & \underline{97.0} & 85.2 & - \\
& INP-Former~\cite{inp_former} & CVPR 2025
& \underline{96.6} & \underline{98.2} & \underline{96.4} & \underline{97.0} & \underline{92.6} & \underline{64.0} & 91.4 & 92.2 & \underline{88.6} & 96.3 & \underline{89.5} & \underline{47.3} \\
\cmidrule(lr){2-15}
& \textbf{ANoCo (Ours)} & CVPR 2026
& \textbf{97.9} & \textbf{99.1} & \textbf{96.9} & \textbf{97.7} & \textbf{95.4} & \textbf{64.9} & \textbf{92.7} & \textbf{93.3} & \textbf{88.8} & \textbf{98.7} & \textbf{94.9} & \textbf{51.9} \\
\midrule

\multirow{7}{*}{2-shot}
& SPADE~\cite{spade} & arXiv 2020
& 82.9 & 91.7 & 91.1 & 92.0 & 85.7 & 44.5 & 79.5 & 82.0 & 80.7 & 95.6 & 84.1 & 35.5 \\
& PatchCore~\cite{patchcore} & CVPR 2022
& 86.3 & 93.8 & 92.0 & 93.3 & 82.3 & 53.0 & 79.9 & 82.8 & 81.7 & 95.4 & 80.5 & 38.0 \\
& WinCLIP~\cite{winclip} & CVPR 2023
& 94.4 & 97.0 & 94.4 & 96.0 & 88.4 & 58.4 & 84.6 & 85.8 & 83.0 & 96.8 & 86.2 & 43.5 \\
& PromptAD~\cite{promptad} & CVPR 2024
& 95.7 & - & - & 96.2 & 88.5 & - & 88.3 & - & - & 97.1 & 85.8 & - \\
& KAG-Prompt~\cite{kag_prompt} & AAAI 2025
& 96.6 & \underline{98.5} & - & 96.5 & 91.1 & - & 92.7 & \underline{94.2} & - & \underline{97.4} & 86.7 & - \\
& INP-Former~\cite{inp_former} & CVPR 2025
& \underline{97.0} & 98.2 & \underline{96.7} & \underline{97.2} & \underline{93.1} & \textbf{65.6} & \textbf{94.6} & \textbf{94.9} & \textbf{90.8} & 97.2 & \underline{91.8} & \underline{50.4} \\
\cmidrule(lr){2-15}
& \textbf{ANoCo (Ours)} & CVPR 2026
& \textbf{98.4} & \textbf{99.2} & \textbf{97.3} & \textbf{98.0} & \textbf{96.0} & \underline{65.3} & \underline{93.3} & 92.9 & \underline{90.3} & \textbf{98.7} & \textbf{94.7} & \textbf{53.0} \\
\midrule

\multirow{7}{*}{4-shot}
& SPADE~\cite{spade} & arXiv 2020
& 84.8 & 92.5 & 91.5 & 92.7 & 87.0 & 46.2 & 81.7 & 83.4 & 82.1 & 96.6 & 87.3 & 43.6 \\
& PatchCore~\cite{patchcore} & CVPR 2022
& 88.8 & 94.5 & 92.6 & 94.3 & 84.3 & 55.0 & 85.3 & 87.5 & 84.3 & 96.8 & 84.9 & 43.9 \\
& WinCLIP~\cite{winclip} & CVPR 2023
& 95.2 & 97.3 & 94.7 & 96.2 & 89.0 & 59.5 & 87.3 & 88.8 & 84.2 & 97.2 & 87.6 & 47.0 \\
& PromptAD~\cite{promptad} & CVPR 2024
& 96.6 & - & - & 96.5 & 90.5 & - & 89.1 & - & - & 97.4 & 86.2 & - \\
& KAG-Prompt~\cite{kag_prompt} & AAAI 2025
& 97.1 & \underline{98.8} & - & 96.7 & 91.4 & - & 93.3 & 94.6 & - & \underline{97.7} & 87.6 & - \\
& INP-Former~\cite{inp_former} & CVPR 2025
& \underline{97.6} & 98.6 & \underline{97.0} & \underline{97.0} & \underline{92.9} & \underline{65.6} & \textbf{96.4} & \textbf{96.0} & \textbf{93.0} & \underline{97.7} & \underline{93.1} & \textbf{54.3} \\
\cmidrule(lr){2-15}
& \textbf{ANoCo (Ours)} & CVPR 2026
& \textbf{98.7} & \textbf{99.3} & \textbf{97.6} & \textbf{98.1} & \textbf{96.2} & \textbf{65.7} & \underline{95.2} & \underline{95.1} & \underline{91.7} & \textbf{98.8} & \textbf{95.7} & \underline{54.2} \\
\bottomrule
\end{tabular}}
\end{table*}

\noindent\textbf{Baselines.}
To demonstrate the effectiveness of our approach, we compare it against a set of recent training-free or normal-only few-shot anomaly detection methods, including SPADE~\cite{spade}, PatchCore~\cite{patchcore}, WinCLIP~\cite{winclip}, PromptAD~\cite{promptad}, KAG-Prompt~\cite{kag_prompt}, and INP-Former~\cite{inp_former}.

\noindent\textbf{Implementation details.}
We adopt the publicly available DINOv3-L/16~\cite{dinov3} backbone as our visual encoder and use the output of its 18-th transformer layer as the middle-level representation. 
All parameters in the backbone and the rest of the framework are frozen, and no additional learnable parameters are introduced.
For more implementation details, please refer to the supplementary material.

\subsection{Main results}
Table~\ref{tab:cr_main} presents a comprehensive comparison of ANoCo against representative few-shot anomaly detection baselines on MVTec-AD and VisA, utilizing six multi-faceted evaluation metrics.
Although recent methods on MVTec-AD already achieve near-saturated performance, ANoCo obtains consistent improvements across the 1-, 2-, and 4-shot settings.
On VisA, ANoCo likewise achieves consistent gains over existing baselines, with particularly notable improvements in localization.

\subsection{Visualization}
\begin{figure*}[t!]
    \centering
    \includegraphics[width=\textwidth]{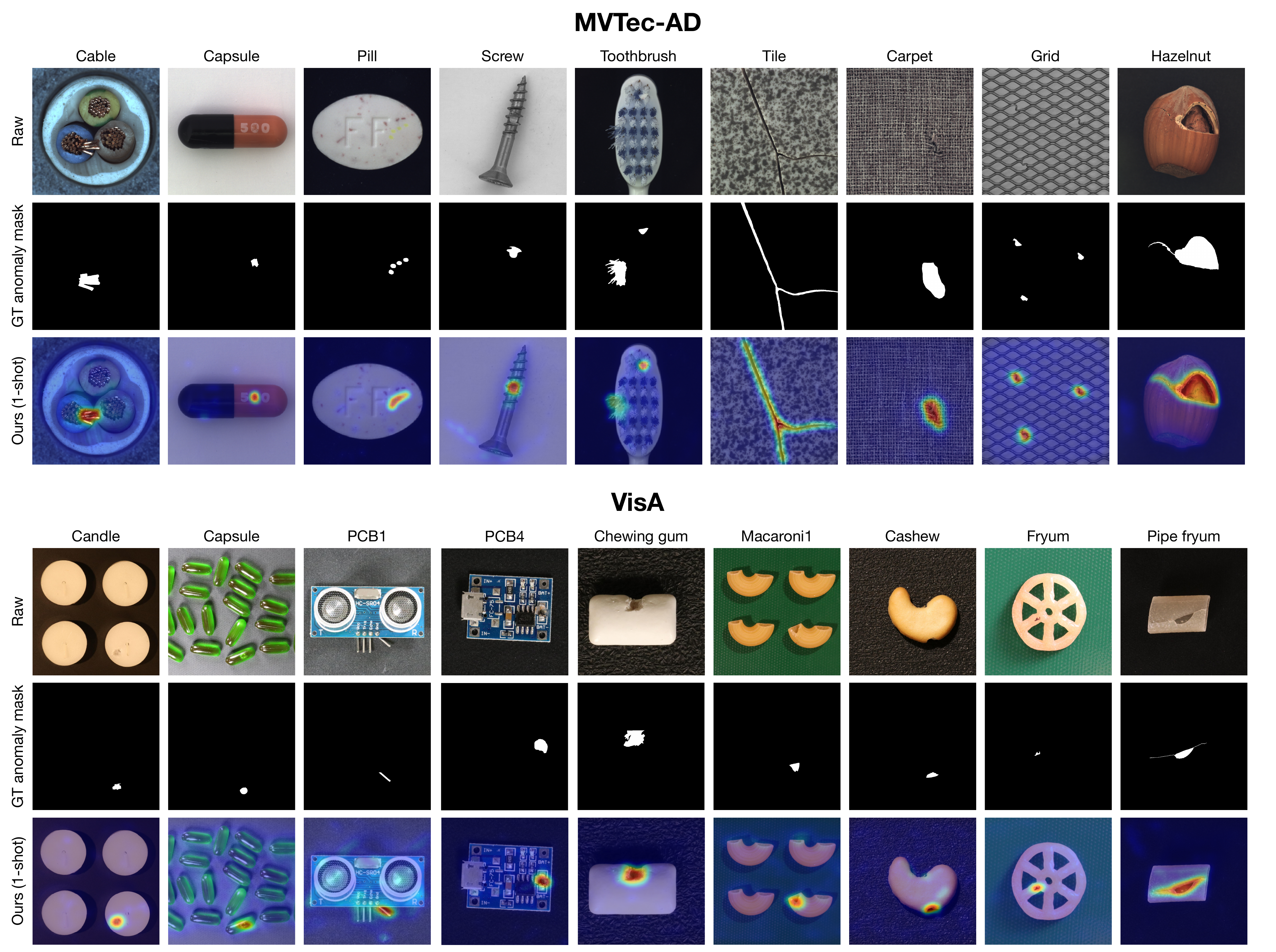}
    \vspace{-0.5cm}
    \caption{\textbf{\mymethod anomaly map visualizations under 1-shot setting.} \mymethod effectively localizes diverse and fine-grained anomalies across the MVTec-AD and VisA.}
\label{fig:vis_2}
\vspace{0.2cm}
\end{figure*}

To demonstrate the anomaly localization capability of ANoCo, we visualize the results obtained under the 1-shot setting across various categories and diverse anomaly types in both the MVTec-AD and VisA datasets in Figure~\ref{fig:vis_2}.
Despite substantial variations in anomaly shape, size, and appearance, ANoCo consistently produces accurate and stable localization results.
These visualizations highlight the robustness and generalization ability of ANoCo, showing that it can effectively identify anomalous regions even when confronted with subtle defects, complex structures, or highly irregular anomaly patterns.

\subsection{Ablation study}

\begin{table*}[t]
\vspace{-0.1cm}
\centering
\caption{\textbf{Component-wise ablation study of ANoCo in the 1-shot setting.} The results demonstrate the individual and collective effectiveness of energy drift formulation, bipartite graph construction, and anchor-driven retrieval.}
\vspace{-0.2cm}
\label{tab:cr_ablation}
\renewcommand{\arraystretch}{1.15}
\setlength{\tabcolsep}{5pt}

\resizebox{\textwidth}{!}{
\begin{tabular}{c ccc ccc ccc ccc}
\toprule
\multirow{3}{*}{\textbf{Method}} & \multicolumn{6}{c}{\textbf{MVTec-AD}} & \multicolumn{6}{c}{\textbf{VisA}} \\
\cmidrule(lr){2-7} \cmidrule(lr){8-13}
& \multicolumn{3}{c}{Image-level} & \multicolumn{3}{c}{Pixel-level} & \multicolumn{3}{c}{Image-level} & \multicolumn{3}{c}{Pixel-level} \\
\cmidrule(lr){2-4} \cmidrule(lr){5-7} \cmidrule(lr){8-10} \cmidrule(lr){11-13}
& AUROC & AUPR & F1-MAX & AUROC & PRO & F1-MAX
& AUROC & AUPR & F1-MAX & AUROC & PRO & F1-MAX \\
\midrule
$k$-NN (L2) & 87.7 & 92.5 & 91.6 & 95.1 & 91.2 & 55.5 & 72.2 & 77.3 & 79.1 & 96.2 & 88.3 & 34.3 \\
$k$-NN (Mahl.) & 93.1 & 96.3 & 94.8 & 94.8 & 92.8 & 56.8 & 77.0 & 81.6 & 87.2 & 95.2 & 94.0 & 34.9 \\
\midrule
$k$-NN + Non-Bip. & 93.7 & 97.2 & 95.2 & 95.8 & 92.2 & 59.3 & 86.3 & 88.3 & 82.9 & 97.5 & 90.2 & 46.6 \\
$k$-NN + Bip. & 95.8 & 98.0 & 96.1 & 97.4 & 94.4 & 62.3 & 90.5 & 91.6 & 86.9 & 98.5 & 94.2 & 50.3 \\
\midrule
\textbf{ANoCo (Anchor-driven + Bip.)} & \textbf{97.9} & \textbf{99.1} & \textbf{96.9} & \textbf{97.7} & \textbf{95.4} & \textbf{64.9} & \textbf{92.7} & \textbf{93.3} & \textbf{88.8} & \textbf{98.7} & \textbf{94.9} & \textbf{51.9} \\
\bottomrule
\end{tabular}
}
\vspace{-0.4cm}
\end{table*}

\noindent\textbf{Contributions of energy drift, bipartite graph, and anchor-driven retrieval.}
Table~\ref{tab:cr_ablation} details the incremental performance gains as we evolve from standard distance measures to our final framework.
Standard $k$-NN baselines under the 5\% setting exhibit the lowest performance because independent similarity scoring fails to detect queries that are locally plausible but globally inconsistent with the normal manifold.
Transitioning to graph energy on a non-bipartite structure begins to model inter-patch relationships, yet performance remains constrained as naive retrieval still introduces structurally incompatible neighbors that weaken manifold coherence.
Incorporating bipartite graph construction further improves performance by preventing anomalies from self-reinforcing through neighborhood consensus and by reducing the oversmoothing effects that typically dilute anomaly evidence.
Finally, switching to anchor-driven retrieval yields the best performance by constructing a representative manifold that is more robust to sparsely sampled multi-modal normality and outliers.
This design effectively preserves fine-grained deviations, leading to more discriminative and precise localization.

\vspace{-0.05cm}
\begin{table*}[t]
  \centering
  \caption{\textbf{Backbone ablation in the 1-shot setting.} ANoCo demonstrates broadly superior performance when evaluated under the same backbone. Resolution is matched per baseline for fair comparison.}
  \label{tab:cr_backbones}
  \vspace{-0.2cm}
  \renewcommand{\arraystretch}{1.10}
  \setlength{\tabcolsep}{4pt}
  \small
  \resizebox{\textwidth}{!}{
    \begin{tabular}{cc ccc ccc ccc ccc}
      \toprule
      \multirow{3}{*}{\textbf{Backbone}} & \multirow{3}{*}{\textbf{Method}} 
      & \multicolumn{6}{c}{\textbf{MVTec-AD}} 
      & \multicolumn{6}{c}{\textbf{VisA}} \\
      \cmidrule(lr){3-8} \cmidrule(lr){9-14}
      & & \multicolumn{3}{c}{Image-level} & \multicolumn{3}{c}{Pixel-level} & \multicolumn{3}{c}{Image-level} & \multicolumn{3}{c}{Pixel-level} \\
      \cmidrule(lr){3-5} \cmidrule(lr){6-8} \cmidrule(lr){9-11} \cmidrule(lr){12-14}
      & & AUROC & AP & F1-MAX & AUROC & PRO & F1-MAX & AUROC & AP & F1-MAX & AUROC & PRO & F1-MAX \\
      \midrule

      \multirow{3}{*}{WideResNet50~\cite{wrn}}
        & SPADE~\cite{spade}
        & 81.0 & 90.6 & 90.3 & 91.2 & 83.9 & 42.4 
        & 81.7 & 83.4 & 82.1 & 96.2 & 85.7 & 40.5 \\
      & PatchCore~\cite{patchcore}
        & 83.4 & 92.2 & 90.5 & 92.0 & 79.7 & 58.4 
        & 81.6 & 84.8 & 82.5 & 96.1 & 82.6 & 41.0 \\
      & \textbf{ANoCo (Ours)}
        & 89.7 & 95.4 & 93.9 & 93.5 & 82.4 & 47.5
        & 83.8 & 87.0 & 82.1 & 96.8 & 80.2 & 42.7 \\

      \midrule

      \multirow{3}{*}{CLIP-B-P~\cite{clip}}
        & WinCLIP~\cite{winclip}
        & 93.1 & 96.5 & 93.7 & 95.2 & 87.1 & 55.9 
        & 83.8 & 85.1 & 83.1 & 96.4 & 85.1 & 41.3 \\
      & PromptAD~\cite{promptad}
        & 94.6 & - & - & 95.9 & 87.9 & - 
        & 86.9 & - & - & 96.7 & 85.8 & - \\
      & \textbf{ANoCo (Ours)}
        & 95.0 & 97.5 & 94.9 & 96.3 & 88.4 & 50.9
        & 87.5 & 89.3 & 84.7 & 97.6 & 83.5 & 37.1 \\

      \midrule

      \multirow{2}{*}{DINOv2-B-R~\cite{dinov2}}
        & INP-Former~\cite{inp_former}
        & 96.6 & 98.2 & 96.4 & 97.0 & 92.6 & 64.0
        & 91.4 & 92.2 & 88.6 & 96.3 & 89.5 & 47.3 \\
      & \textbf{ANoCo (Ours)}
        & 97.3 & 98.8 & 96.9 & 97.1 & 93.0 & 62.3
        & 91.8 & 92.1 & 88.6 & 98.2 & 93.6 & 47.2 \\
        
      \midrule

      \multirow{3}{*}{DINOv3-L~\cite{dinov3}}
      & $k$-NN (Full-shot)
        & 83.7 & 94.0 & 91.2 & 84.0 & 72.8 & 44.3
        & 83.6 & 86.9 & 81.8 & 96.0 & 84.5 & 40.4 \\
      & \textbf{ANoCo (1-shot)}
        & 97.9 & 99.1 & 96.9 & 97.7 & 95.4 & 64.9
        & 92.7 & 93.3 & 88.8 & 98.7 & 94.9 & 51.9 \\
      & \textbf{ANoCo (Full-shot)}
        & 99.4 & 99.7 & 98.5 & 98.2 & 96.4 & 67.8
        & 98.7 & 98.9 & 96.1 & 99.2 & 95.3 & 54.7 \\
      \bottomrule
    \end{tabular}
  }
\vspace{-0.3cm}
\end{table*}
\noindent\textbf{Backbone fairness.}
For better comparison, we compare against each baseline under its corresponding backbone and input resolution in Table~\ref{tab:cr_backbones}.
On most metrics, ANoCo consistently outperforms both training-free and training-based baselines across all evaluated backbones and datasets.
This suggests that the performance gains of ANoCo are not merely a result of feature quality, but are driven by our energy-based non-conformity formulation which effectively captures structural violations of the normal manifold.

\begin{table}[t]
  \centering
    \caption{\textbf{Few-shot performance on the OCT2017 dataset}. ANoCo consistently outperforms the baselines, suggesting its applicability beyond industrial benchmarks.}
    \vspace{-0.1cm}
  \label{tab:cr_benchmarks}
  \renewcommand{\arraystretch}{1.0}
  \setlength{\tabcolsep}{8pt}
  \resizebox{0.7\columnwidth}{!}{%
    \begin{tabular}{cc cc}
      \toprule
      \multirow{2}{*}{\textbf{\# refs}} & \multirow{2}{*}{\textbf{Method}} & \multicolumn{2}{c}{\textbf{OCT2017}} \\
      \cmidrule(lr){3-4}
      & & AUROC & AP \\
      \midrule
      
      \multirow{6}{*}{2-shot} 
        & PatchCore~\cite{patchcore} & 73.0 & 86.6 \\
        & RegAD~\cite{regad}         & 70.1 & 84.6 \\
        & WinCLIP~\cite{winclip}     & 94.7 & 98.3 \\
        & InCTRL~\cite{inctrl}       & 94.9 & 98.3 \\
        \cmidrule{2-4}
        & \textbf{ANoCo (WRN)}       & 95.0 & 97.8 \\
        & \textbf{ANoCo (CLIP)}      & \textbf{97.7} & \textbf{99.2} \\
      \midrule
      
      \multirow{6}{*}{4-shot} 
        & PatchCore~\cite{patchcore} & 76.8 & 88.1 \\
        & RegAD~\cite{regad}         & 72.7 & 86.1 \\
        & WinCLIP~\cite{winclip}     & 96.2 & 98.6 \\
        & InCTRL~\cite{inctrl}       & 96.8 & 98.9 \\
        \cmidrule{2-4}
        & \textbf{ANoCo (WRN)}       & 96.6 & 98.7 \\
        & \textbf{ANoCo (CLIP)}      & \textbf{99.4} & \textbf{99.7} \\
      \midrule
      
      \multirow{6}{*}{8-shot} 
        & PatchCore~\cite{patchcore} & 80.6 & 90.4 \\
        & RegAD~\cite{regad}         & 74.4 & 88.6 \\
        & WinCLIP~\cite{winclip}     & 97.0 & 99.0 \\
        & InCTRL~\cite{inctrl}       & 97.4 & 99.1 \\
        \cmidrule{2-4}
        & \textbf{ANoCo (WRN)}       & 97.7 & 99.1 \\
        & \textbf{ANoCo (CLIP)}      & \textbf{99.3} & \textbf{99.7} \\
      \bottomrule
    \end{tabular}%
  }
\vspace{-0.4cm}
\end{table}
\noindent\textbf{Additional benchmarks.}
In Table~\ref{tab:cr_benchmarks}, we report few-shot results on OCT2017~\cite{oct2017} using the same backbone and resolution as the few-shot baselines (RegAD~\cite{regad}, WinCLIP~\cite{winclip}, InCTRL~\cite{inctrl}), where ANoCo consistently outperforms the baselines.
This result suggests that the applicability of ANoCo may not be limited to representative industrial anomaly detection benchmarks.

\begin{table}[t]
\centering
\caption{\textbf{Forward latency (ms) under different shot settings}. For each backbone, we report the baseline, ANoCo, and the latency breakdown of ANoCo into anchor-driven retrieval and graph optimization.}
\vspace{-0.1cm}
\setlength{\tabcolsep}{2.0mm}

\resizebox{\columnwidth}{!}{%
\renewcommand{\arraystretch}{1.1}
\begin{tabular}{c|cccc}
\toprule
\multicolumn{5}{c}{\textbf{WideResNet50}~\cite{wrn}} \\
\midrule
\textbf{\# refs} & \textbf{PatchCore}~\cite{patchcore} & \textbf{ANoCo} & \textbf{Anchor-driven} & \textbf{Graph opt.} \\
\midrule
1-shot & 7.5 & 9.7 & 1.4 & 0.5 \\
2-shot & 7.6 & 10.3 & 1.5 & 0.5 \\
4-shot & 8.1 & 10.1 & 2.0 & 0.5 \\
\bottomrule
\end{tabular}%
}

\vspace{0.5em}

\resizebox{\columnwidth}{!}{%
\renewcommand{\arraystretch}{1.10}
\begin{tabular}{c|cccc}
\toprule
\multicolumn{5}{c}{\textbf{CLIP-B-P}~\cite{clip}} \\
\midrule
\textbf{\# refs} & \textbf{WinCLIP}~\cite{winclip} & \textbf{ANoCo} & \textbf{Anchor-driven} & \textbf{Graph opt.} \\
\midrule
1-shot & 1560.0 & 14.0 & 1.3 & 0.5 \\
2-shot & 1615.0 & 14.2 & 1.3 & 0.5 \\
4-shot & 1803.6 & 13.8 & 1.3 & 0.5 \\
\bottomrule
\end{tabular}%
}
\label{tab:cr_latency}
\vspace{-0.2cm}
\end{table}
\noindent\textbf{Latency breakdown.}
We report per-sample latency under matched backbones and identical hardware settings in Table~\ref{tab:cr_latency}.
The results show that the closed-form optimization in ANoCo is practical, while remaining substantially more efficient than WinCLIP.
\section{Conclusion}
We introduced \textbf{ANoCo}, a training-free, energy-based framework that interprets anomaly as non-conformity to a fixed normal manifold rather than low similarity to individual exemplars.
By constructing a bipartite query–reference graph with anchor-driven retrieval and formulating anomaly scoring as anchored Laplacian energy minimization with a closed-form solution, ANoCo avoids evidence-diluting query–query and reference–reference interactions and instead measures how much each query patch must move to become compatible with normal data; the resulting feature-drift energy provides a physically interpretable anomaly score and stable, fine-grained anomaly maps.
Extensive experiments on MVTec-AD and VisA in the few-shot regime show that ANoCo consistently improves image-level detection and pixel-level localization over baselines while remaining simple to deploy, requiring no additional training.
\clearpage
\section*{Acknowledgments} This work was supported by Samsung Electronics Global Technology Research, the Institute of Information and communications Technology Planning and evaluation (IITP) grant (No. RS-2025-25422680, No. RS-2025-25441560, No. RS-2020-II201373), and the National Research Foundation of Korea (NRF) grant funded by the Korea government (MSIT) (No. RS-2025-24533064).  
{
    \small
    \bibliographystyle{ieeenat_fullname}
    \bibliography{main}
}

\appendix

\setcounter{section}{0}
\setcounter{figure}{0}
\setcounter{table}{0}
\setcounter{equation}{0}

\renewcommand{\thesection}{S\arabic{section}}
\renewcommand{\thefigure}{S\arabic{figure}}
\renewcommand{\thetable}{S\arabic{table}}
\renewcommand{\theequation}{S\arabic{equation}}

\clearpage
\setcounter{page}{1}
\maketitlesupplementary
\section*{Table of Contents}
\color{black}

\begin{description}[
    leftmargin=3em,
    labelwidth=2.5em,
    labelsep=0.5em,
    itemsep=5pt,   
    topsep=2pt,    
    parsep=0pt
]
    \item[\textbf{S1.}] Implementation details \dotfill \hyperref[sec:implementation]{11}
    \item[\textbf{S2.}] Ablation on the query feature stabilization coefficient $\Lambda_q$ \dotfill \hyperref[sec:lambda]{11}
    \item[\textbf{S3.}] Graph update strategies \dotfill \hyperref[sec:graph]{11}
    \item[\textbf{S4.}] Additional details of non-conformity as anomaly evidence \dotfill \hyperref[sec:score]{12}
    \item[\textbf{S5.}] Evaluation on Real-IAD dataset \dotfill \hyperref[sec:realiad]{12}
    \item[\textbf{S6.}] Additional results \dotfill \hyperref[sec:additional]{13}
    \item[\textbf{S7.}] Per-category performance and detailed visualization \dotfill \hyperref[sec:category]{13}
\end{description}

\section{Implementation details}
\label{sec:implementation}
\noindent \textbf{Backbone, preprocessing, and post-processing.}
We use DINOv3-L/16~\cite{dinov3} as a frozen visual backbone and extract patch-level features from the 18-th Transformer block.
All input images are resized to $768 \times 768$ and normalized with mean $(0.485, 0.456, 0.406)$ and standard deviation $(0.229, 0.224, 0.225)$.
The resulting anomaly map is further smoothed with a Gaussian filter with kernel size $7$ and $\sigma = 0.8$.

\noindent \textbf{Inference-time augmentation.}
At test time, we apply geometric augmentation to both the query image and the few-shot reference images. For each reference image, we retain the original view and generate additional transformed views using random horizontal flipping, random vertical flipping, and an affine transformation consisting of a rotation sampled from $[-15^\circ, -5^\circ] \cup [5^\circ, 15^\circ]$, a translation within $\pm 2\%$ of the image width and height, isotropic scaling in $[0.95, 1.05]$, and shear sampled from $[-5^\circ, 5^\circ]$ along both axes. We generate 25 paired augmented views and 5 additional reference-only augmented views for each reference image. For the paired views, the same transformation parameters are applied to both the query image and the corresponding reference image, whereas the reference-only views are used only to enrich the reference patch pool. During inference, anomaly maps from the augmented query views are mapped back to the original image coordinates and aggregated using entropy-based confidence weighting.

\noindent \textbf{Memory usage.}
 In the 1-shot setting on a single NVIDIA RTX 3090 GPU, the maximum memory usage was approximately 10 GB when augmentation was enabled and 2 GB when augmentation was disabled.

\section{Ablation on the query feature stabilization coefficient $\Lambda_q$}
\label{sec:lambda}
\begin{figure}[t!]
    \centering
    \includegraphics[width=\columnwidth]{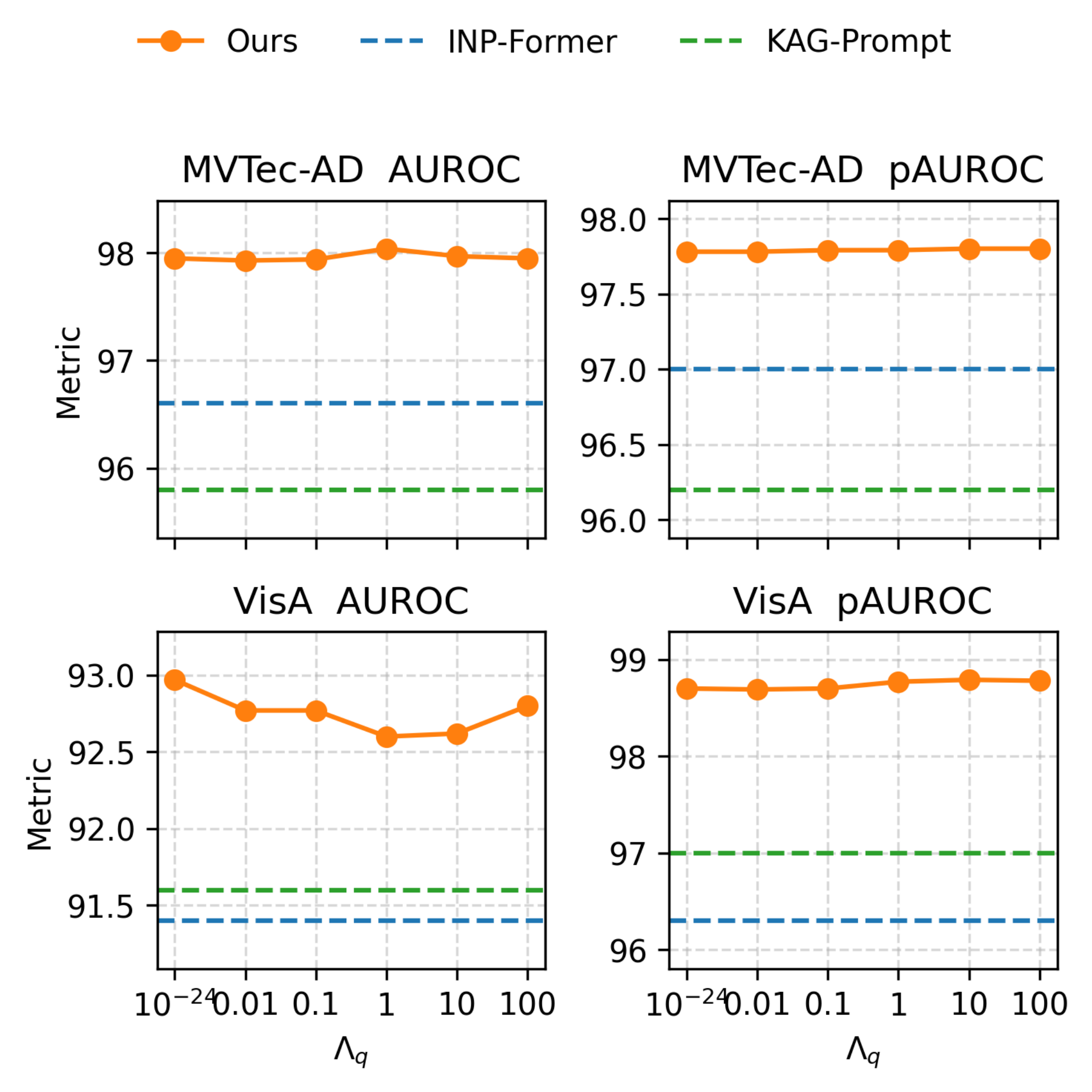}
    \caption{\textbf{Performance trend of ANoCo with respect to $\Lambda_q$ in the 1-shot setting.}}
\label{fig:lambda_q}
\end{figure}
\noindent We analyze the effect of the query feature stabilization coefficient $\Lambda_q$ in Figure~\ref{fig:lambda_q}. 
We sweep $\Lambda_q \in \{10^{-24}, 0.01, 0.1, 1, 10, 100\}$ and observe that ANoCo consistently outperforms all baselines across all values of $\Lambda_q$ on every dataset. 
The overall performance curves are relatively flat, indicating that ANoCo is robust to the exact choice of $\Lambda_q$ and does not require delicate tuning of this coefficient. 
At the same time, some datasets exhibit small additional gains for moderate values of $\Lambda_q$ (e.g., around $\Lambda_q$ = 1), suggesting that query feature stabilization can be exploited as an optional degree of freedom when further fine-tuning is desired.
\begin{figure*}[t!]
    \centering
    \includegraphics[width=0.7\textwidth]{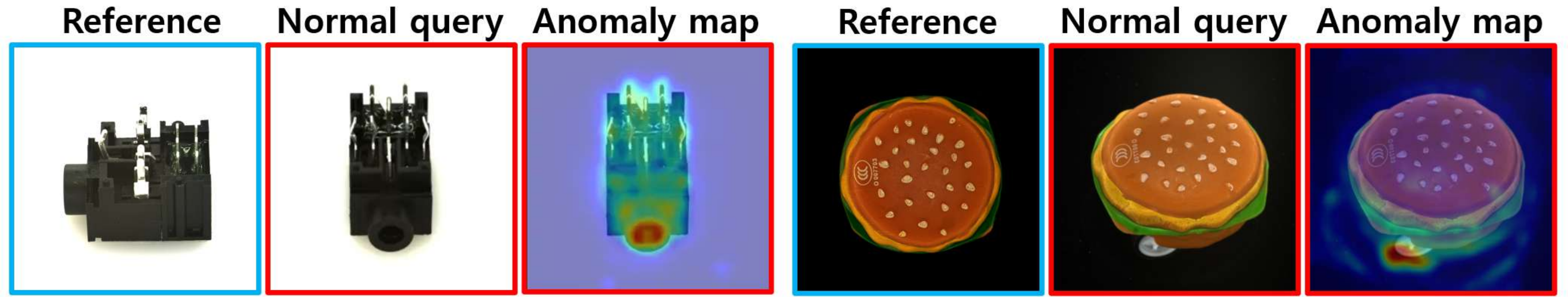}
    \caption{\textbf{Failure cases on Real-IAD in the 1-shot setting.}}
\label{fig:realiad_fail}
\end{figure*}
\section{Graph update strategies}
\label{sec:graph}
\begin{table}[t]
\vspace{0.2cm}
\centering
\small
\renewcommand{\arraystretch}{1.2}
\setlength{\tabcolsep}{6pt}
\caption{\textbf{Ablation study on graph update strategies under the 1-shot setting}. We compare message passing with our energy minimization approach.}
\resizebox{0.9\columnwidth}{!}{
\begin{tabular}{c cc cc}
\toprule
\multirow{2}{*}{\textbf{Round}}
& \multicolumn{2}{c}{\textbf{MVTec-AD}}
& \multicolumn{2}{c}{\textbf{VisA}} \\
\cmidrule(lr){2-3} \cmidrule(lr){4-5}
& AUROC & pAUROC
& AUROC & pAUROC \\
\midrule
1  & 87.7 & 93.7 & 67.5 & 95.3 \\
2  & 82.7 & 92.2 & 65.1 & 93.3 \\
3  & 82.4 & 92.2 & 65.6 & 93.3 \\
4  & 82.4 & 92.2 & 65.3 & 93.3 \\
5  & 82.4 & 92.2 & 65.6 & 93.2 \\
\midrule
\textbf{ANoCo (Ours)} & \textbf{97.9} & \textbf{97.7} & \textbf{92.7} & \textbf{98.7} \\
\bottomrule
\end{tabular}
}
\label{tab:cr_message_passing}
\end{table}
Table~\ref{tab:cr_message_passing} presents an ablation study on graph update strategies under the 1-shot setting.
We compare the proposed query optimization via energy minimization with a training-free message-passing alternative.
For message passing, we perform iterative propagation using cosine-similarity-based edge weights, which are identical to those used in the energy-minimization formulation.

We evaluate message passing with 1 to 5 propagation rounds.
As the number of rounds increases, message passing does not yield consistent improvements in anomaly discrimination, making the choice of the optimal number of rounds non-trivial.
In contrast, the energy-minimization strategy of ANoCo achieves consistently stronger and more stable performance across both datasets than the message-passing variants.

\begin{table}[h!]
\centering
\small
\renewcommand{\arraystretch}{1.2}
\setlength{\tabcolsep}{6pt}
\caption{Ablation study of various ANoCo non-conformity metrics under the 1-shot setting.}
\resizebox{0.9\columnwidth}{!}{
\begin{tabular}{c cc cc}
\toprule
\multirow{2}{*}{\textbf{Metric}}
& \multicolumn{2}{c}{\textbf{MVTec-AD}}
& \multicolumn{2}{c}{\textbf{VisA}} \\
\cmidrule(lr){2-3} \cmidrule(lr){4-5}
& AUROC & pAUROC
& AUROC & pAUROC \\
\midrule
L2              & 97.6 & 97.6 & 92.7 & 98.7 \\
Cos-dis         & 97.7 & 97.6 & 92.4 & 98.6 \\
Eq.~\textcolor{cvprblue}{10} & 97.9 & 97.7 & 92.7 & 98.7 \\
\bottomrule
\end{tabular}
}
\label{tab:cr_anomaly_score}
\end{table}
\section{Additional details of non-conformity as anomaly evidence}
\label{sec:score}
We additionally compare three candidates for the patchwise non-conformity energy $E_i$ used as anomaly evidence: pure Euclidean energy $\lVert \tilde{\mathbf{f}}^{q}_i - \mathbf{f}^{q}_i \rVert_2^2$, pure cosine dissimilarity $(1 - \cos(\tilde{\mathbf{f}}^{q}_i, \mathbf{f}^{q}_i))$, and their product $\lVert \tilde{\mathbf{f}}^{q}_i - \mathbf{f}^{q}_i \rVert_2^2 (1 - \cos(\tilde{\mathbf{f}}^{q}_i, \mathbf{f}^{q}_i))$ as adopted in Equation \textcolor{cvprblue}{10} in Table~\ref{tab:cr_anomaly_score}.
All three options consistently outperform the baselines, and the performance differences between them are relatively small. 
However, the product form provides a small but consistent improvement over using either term alone. This suggests that ANoCo does not rely on a carefully tuned or dataset-specific non-conformity formulation, while the chosen energy yields a slightly stronger separation between normal and anomalous regions.

\begin{table}[h!]
\centering
\setlength{\tabcolsep}{4pt}
\renewcommand{\arraystretch}{1.2}
\caption{\textbf{Comparison of anomaly detection and localization performance on Real-IAD dataset under different few-shot settings.}}
\begin{tabular}{c c c c}
\toprule
\multirow{2}{*}{\makecell[c]{\textbf{\# refs}}} & \multirow{2}{*}{\makecell[c]{\textbf{Method}}} 
& \multicolumn{2}{c}{\textbf{Real-IAD}~\cite{realiad}} \\
\cmidrule(lr){3-4}
& & AUROC & pAUROC \\
\midrule

\multirow{6}{*}{\makecell[c]{1-shot}}
& SPADE~\cite{spade}
& 51.2 & 59.5 \\
& PatchCore~\cite{patchcore}
& 59.3 & 89.6 \\
& WinCLIP~\cite{winclip}
& \textbf{69.4} & 91.9 \\
& PromptAD~\cite{promptad}
& 52.2 & 84.9 \\
& INP-Former~\cite{inp_former}
& 67.5 & \textbf{94.9} \\
\cmidrule(lr){2-4}
& \textbf{ANoCo (Ours)}  
& 65.4 & 93.4 \\
\midrule

\multirow{6}{*}{\makecell[c]{2-shot}}
& SPADE~\cite{spade}
& 50.9 & 59.5 \\
& PatchCore~\cite{patchcore}
& 63.3 & 92.0 \\
& WinCLIP~\cite{winclip}
& 70.9 & 93.2 \\
& PromptAD~\cite{promptad}
& 57.7 & 86.4 \\
& INP-Former~\cite{inp_former}
& 70.6 & 96.0 \\
\cmidrule(lr){2-4}
& \textbf{ANoCo (Ours)}  
& \textbf{77.3} & \textbf{97.1} \\
\midrule

\multirow{6}{*}{\makecell[c]{4-shot}}
& SPADE~\cite{spade}
& 50.8 & 59.5 \\
& PatchCore~\cite{patchcore}
& 66.0 & 92.9 \\
& WinCLIP~\cite{winclip}
& 73.0 & 93.8 \\
& PromptAD~\cite{promptad}
& 59.7 & 86.9 \\
& INP-Former~\cite{inp_former}
& 76.7 & 97.3 \\
\cmidrule(lr){2-4}
& \textbf{ANoCo (Ours)}  
& \textbf{83.8} & \textbf{98.3} \\
\bottomrule
\end{tabular}
\label{tab:realiad}
\end{table}
\section{Evaluation on Real-IAD dataset}
\label{sec:realiad}

In this section, we evaluate ANoCo on the Real-IAD~\cite{realiad} dataset, and report the results in Table~\ref{tab:realiad}, complementing the MVTec-AD and VisA results in the main paper.
Real-IAD consists of 30 object categories, with 36,645 normal images for training and a test set of 63,256 normal images and 51,329 anomalous images.
ANoCo shows strong performance in the 2-shot and 4-shot settings, significantly outperforming the baseline.
We also analyze failure cases and observe that, in the 1-shot setting, ANoCo can misclassify previously unseen but normal variations as anomalies. This issue is particularly pronounced on Real-IAD due to the large viewpoint variation in query images (Figure~\ref{fig:realiad_fail}).
However, with more reference images, ANoCo can better capture the normal manifold, resulting in strong performance in the 2-shot and 4-shot settings.

\begin{figure*}[t!]
    \centering
    \includegraphics[width=\textwidth]{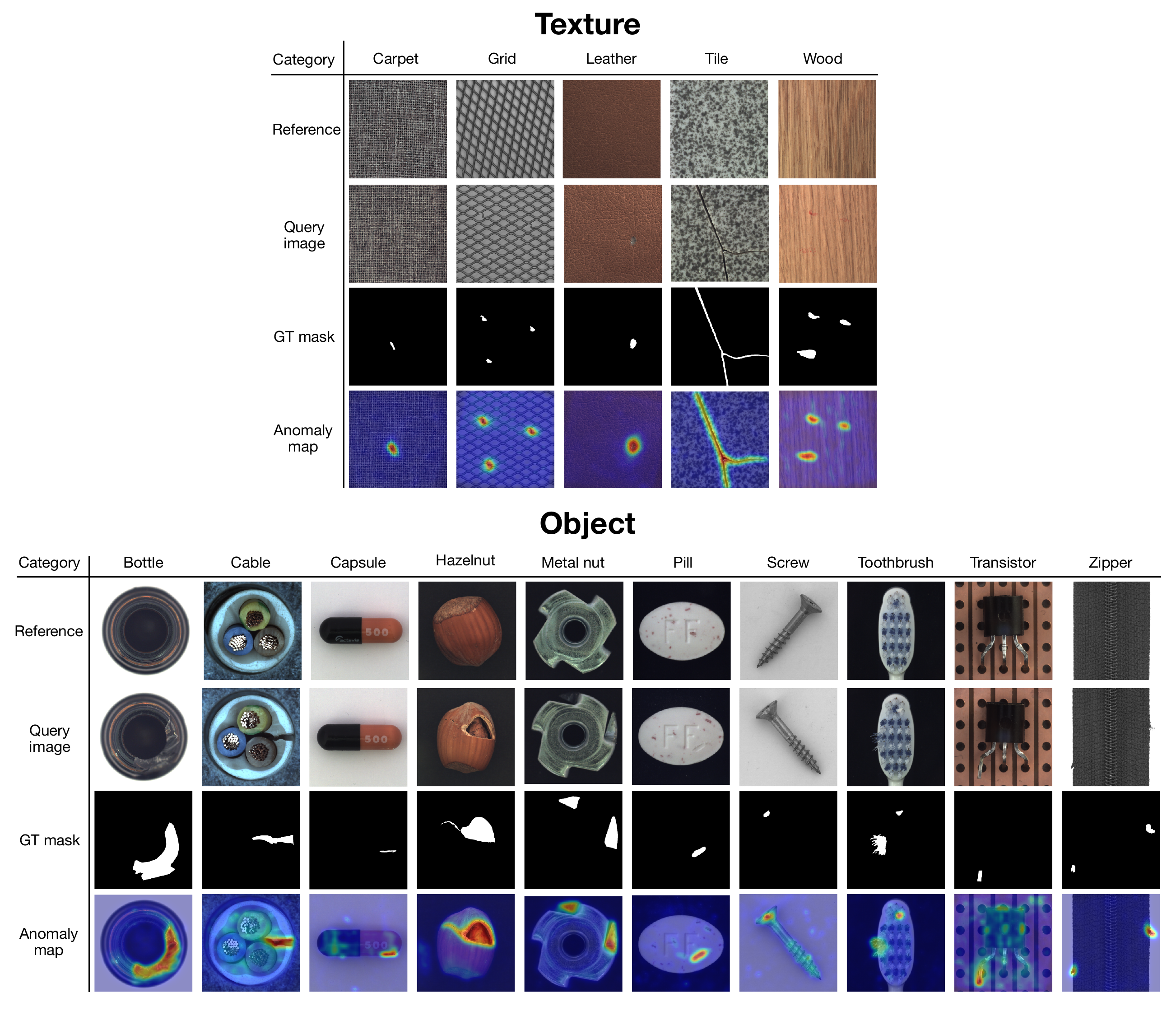}
    \caption{\textbf{Qualitative visualization of anomaly localization results on MVTec-AD under the 1-shot setting.}}
\label{fig:cr_vis}
\end{figure*}

\begin{table}[h]
\centering
\caption{\textbf{Pixel-level AUROC comparison with AD-DINOv3.}}
\label{tab:cr_ad-dinov3}
\begin{tabular}{ccc}
\toprule
\textbf{Method} & \textbf{MVTec-AD} & \textbf{VisA} \\
\midrule
AD-DINOv3~\cite{ad-dinov3} & 91.6 & 95.6 \\
\textbf{ANoCo (Ours)}   & \textbf{97.7} & \textbf{98.7} \\
\bottomrule
\end{tabular}
\end{table}
\begin{table}[h]
\centering
\caption{\textbf{Preliminary generality analysis with WinCLIP on MVTec-AD under the 1-shot setting.}}
\label{tab:cr_generality}
\renewcommand{\arraystretch}{1.2}
\setlength{\tabcolsep}{10pt}
\resizebox{0.9\columnwidth}{!}{%
\begin{tabular}{c cc}
\toprule
\multirow{2}{*}{\textbf{Method}} & \multicolumn{2}{c}{\textbf{MVTec-AD}} \\
\cmidrule(lr){2-3}
 & AUROC & AP \\
\midrule
WinCLIP      & 93.1 & 96.5 \\
WinCLIP + \textbf{ANoCo (Ours)}  & \textbf{93.8} & \textbf{97.1} \\
\bottomrule
\end{tabular}%
}
\end{table}

\section{Additional results}
\label{sec:additional}
\noindent \textbf{Generality.}
ANoCo is training-free and can be used with different visual feature extractors, but it is not intended as a fully plug-and-play post-processing framework.
Rather, it serves as an alternative to retrieval- or similarity-based few-shot anomaly detection methods.
As a preliminary observation, Table~\ref{tab:cr_generality} shows that the proposed graph optimization module can also be applied to another baseline, WinCLIP.

\noindent \textbf{Comparison with AD-DINOv3.}
AD-DINOv3~\cite{ad-dinov3} uses both DINOv3 and CLIP and requires training on both anomalous and normal samples from external datasets. In this work, we focus on a setting where only few-shot normal samples within the same dataset are available. Regardless, we compare against the reported results of AD-DINOv3, where ANoCo is shown to outperform in Table~\ref{tab:cr_ad-dinov3}.

\begin{table}[!t]
\centering
\caption{\textbf{Per-category anomaly detection and localization performance of ANoCo on MVTec-AD in the 1-shot setting.}}
\label{tab:cr_category}
\setlength{\tabcolsep}{3pt}
\renewcommand{\arraystretch}{1.0}

\resizebox{\columnwidth}{!}{
\begin{tabular}{c ccc ccc}
\toprule
\multirow{2}{*}{\textbf{Category}}
& \multicolumn{3}{c}{\textbf{Image-level}}
& \multicolumn{3}{c}{\textbf{Pixel-level}} \\
\cmidrule(lr){2-4} \cmidrule(lr){5-7}
& AUROC & AUPR & F1-MAX
& AUROC & PRO & F1-MAX \\
\midrule
\multicolumn{7}{c}{\textit{\textbf{Texture}}} \\
\midrule
carpet     & 99.8 & 99.9 & 100.0 & 99.6 & 98.7 & 76.9 \\
grid       & 100.0 & 100.0 & 100.0 & 99.5 & 96.7 & 60.3 \\
leather    & 100.0 & 100.0 & 100.0 & 99.3 & 98.2 & 46.5 \\
tile       & 100.0 & 100.0 & 100.0 & 98.1 & 95.8 & 75.1 \\
wood       & 98.6 & 99.5 & 97.5 & 97.0 & 96.1 & 70.4 \\
\midrule
Mean       & 99.6 & 99.8 & 99.5 & 98.7 & 97.1 & 65.8 \\
\midrule
\multicolumn{7}{c}{\textit{\textbf{Object}}} \\
\midrule
bottle     & 99.9 & 99.9 & 99.2 & 98.7 & 97.0 & 78.1 \\
cable      & 96.0 & 97.9 & 93.6 & 97.0 & 93.6 & 72.3 \\
capsule    & 93.8 & 98.6 & 93.9 & 98.8 & 97.4 & 57.5 \\
hazelnut   & 98.2 & 99.0 & 97.1 & 99.6 & 95.5 & 81.2 \\
metal nut & 100.0 & 100.0 & 100.0 & 95.5 & 94.8 & 70.1 \\
pill       & 96.5 & 99.2 & 97.5 & 95.0 & 97.8 & 47.9 \\
screw      & 89.7 & 96.8 & 88.5 & 98.4 & 93.5 & 53.2 \\
toothbrush & 100.0 & 100.0 & 100.0 & 99.3 & 97.2 & 66.2 \\
transistor & 96.7 & 96.1 & 88.6 & 90.9 & 80.6 & 48.6 \\
zipper     & 99.4 & 99.8 & 98.7 & 99.3 & 97.6 & 68.8 \\
\midrule
Mean       & 97.0 & 98.7 & 95.7 & 97.2 & 94.5 & 64.3 \\
\midrule
\midrule
Overall Mean & 97.9 & 99.1 & 96.9 & 97.7 & 95.4 & 64.9 \\
\bottomrule
\end{tabular}
}

\vspace{-0.2cm}
\end{table}
\section{Per-category performance and detailed visualization}
\label{sec:category}
Table~\ref{tab:cr_category} reports per-category anomaly detection and localization results on MVTec-AD under the 1-shot setting, providing a more fine-grained view beyond the dataset-level averages reported in the main paper. We also include detailed visualizations to further illustrate category-wise behavior and qualitative characteristics.

\end{document}